\documentclass[journal]{IEEEtran}
\ifCLASSINFOpdf
\else
\fi
\usepackage{graphicx}
\graphicspath{{figures/}}
\usepackage{xcolor}

\newcommand{\Eq}[1]{Eq.~(\ref{eq:#1})}
\newcommand{\eq}[1]{\Eq{#1}}
\newcommand{\fig}[1]{Fig.~\ref{fig:#1}}
\newcommand{\tab}[1]{Tab.~\ref{tab:#1}}

\newcommand{\sect}[1]{Section~\ref{sec:#1}}

\usepackage{graphicx}
\usepackage{amsmath}
\usepackage{amssymb}
\usepackage{booktabs}
\usepackage{comment}

\usepackage{xspace}
  \newcommand{\latinphrase}[1]{\textit{#1}}  
\newcommand{\etal}{\latinphrase{et~al.}\xspace}
\newcommand{\ie}{\latinphrase{i.e.}\xspace}

\newcommand{\model}{N-Jet\xspace}
\newcommand{\safespl}{safe-subsampling\xspace}
\newcommand{\Safespl}{Safe-subsampling\xspace}
\newcommand{\layer}{N-Jet convolutional\xspace}

\usepackage{amsmath,amssymb} 
\usepackage{color}
\usepackage{multirow}

\begin{document}
\makeatletter

\def\mycopyrightnotice{%
{1941-0042~\copyright~2021 IEEE\hfill}
\gdef\mycopyrightnotice{}
}
\def\ps@IEEEtitlepagestyle{%
\def\@oddfoot{
\begin{tabular}{ll} 
    \mycopyrightnotice & Personal use is permitted, but republication/redistribution requires IEEE permission.\\
    & See https://www.ieee.org/publications/rights/index.html for more information.\\
\end{tabular}
}
\def\@evenfoot{}%
}
%
\title{Resolution learning in deep convolutional\\ networks using scale-space theory}



%
%
%
\author{Silvia~L.~Pintea$^*$,
    Nergis~T\"{o}men$^*$,
    Stanley~F.~Goes,
    Marco~Loog,
    and Jan~C.~van~Gemert
\thanks{Dr. S.L. Pintea, Dr. N. T\"{o}men, Dr. M. Loog and Dr. J.C. van Gemert are with the Computer Vision Lab, Delft University of Technology, 2628CD Delft, Netherlands (e-mail: s.l.pintea@tudelft.nl; n.tomen@tudelft.nl). MSc. S.F. Goes is with Q.E.F Electronic Innovations.}
\thanks{$^*$ Shared first authorship with equal contributions.}}

\maketitle
\begin{abstract}
Resolution in deep convolutional neural networks (CNNs) is typically bounded by the receptive field size through filter sizes, and subsampling layers or strided convolutions on feature maps. 
The optimal resolution may vary significantly depending on the dataset. 
Modern CNNs hard-code their resolution hyper-parameters in the network architecture which makes tuning such hyper-parameters cumbersome. 
We propose to do away with hard-coded resolution hyper-parameters and aim to learn the appropriate resolution from data. 
We use scale-space theory to obtain a self-similar parametrization of filters and make use of the N-Jet: a truncated Taylor series to approximate a filter by a learned combination of Gaussian derivative filters. 
The parameter $\sigma$ of the Gaussian basis controls both the amount of detail the filter encodes and the spatial extent of the filter. 
Since $\sigma$ is a continuous parameter, we can optimize it with respect to the loss. 
The proposed \model layer achieves comparable performance when used in state-of-the art architectures, while learning the correct resolution in each layer automatically.
We evaluate our \model layer on both classification and segmentation, and we show that learning $\sigma$ is especially beneficial when dealing with inputs at multiple sizes.
\end{abstract}


\IEEEpeerreviewmaketitle

\section{Introduction}
\begin{figure}[t]
\centering
\includegraphics[width=\linewidth]{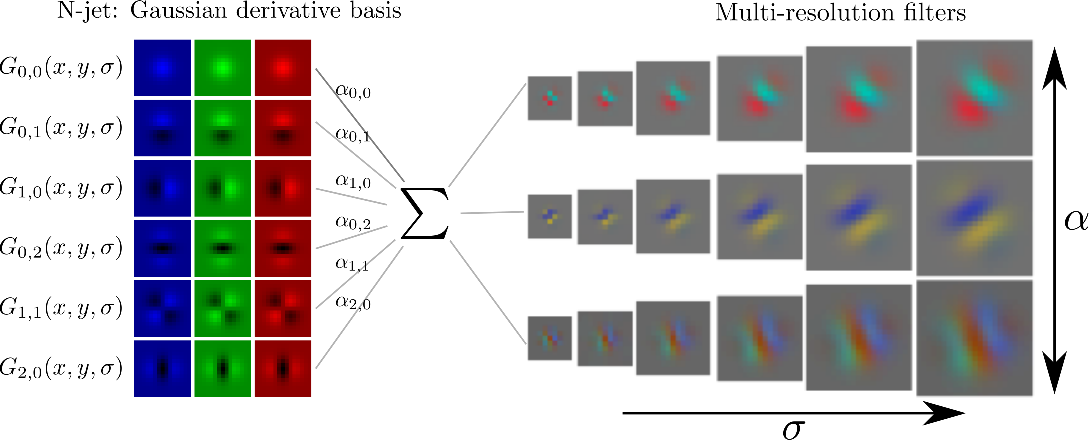} \\
\caption{
    Illustration of how an N-Jet Gaussian derivative basis parametrizes the shape and size of the filters. 
    A linear combination of Gaussian derivative basis filters \textit{(left)} weighted by $\alpha$ parameters span a Taylor series to locally approximate the shape of image filters. 
    The filters are self-similar: the $\sigma$ parameter can change the size of the filters while keeping its spatial structure intact. 
    Each of the three filters \textit{(right)} has a different weighted combination of basis filters, while their $\sigma$ is varied on the horizontal axis. 
    Optimizing for $\alpha$ learns filters shape, optimizing for $\sigma$ learns their size from the data.}
\label{fig:fig1}
\end{figure}
Resolution defines the \emph{inner scale} at which objects should be observed in an image \cite{koenderinkBioCyb84structOfIm}.
To control the resolution in a network, one can change the filter sizes or feature map sizes. 
Because there is a maximum frequency that can be encoded in a limited spatial extent, the filter sizes and feature map sizes define a lower bound on the resolution encoded in the network. 
CNNs typically use small filters of $3 \times 3$ px or $5 \times 5$ px, where the first layers are forced to look at detailed, local image neighborhoods such as edges, blobs, and corners. 
As the network deepens, each subsequent convolution increases the receptive field size linearly~\cite{luoNIPS16understandingRecField}, allowing the network to combine the detailed responses of the previous layer to obtain textures, and object parts. 
Going even deeper, strategically placed memory-efficient subsampling operations reduce feature maps to half their size which is equivalent to increasing the receptive field multiplicatively. 
At the deepest layers, the receptive field spans a large portion of the image and objects emerge as combinations of their parts~\cite{zeiler2014visualizing}. 
The resolution, as controlled by the sizes of the receptive field and feature maps, is one of the fundamental aspects of CNNs.

In modern CNN architectures~\cite{huang2017denseNet,xie2017aggregatedResidual}, the resolution is a hyper-parameter which has to be manually tuned using expert knowledge, by changing the filter sizes or the subsampling layers. 
For example, the popular ResNeXt~\cite{xie2017aggregatedResidual} for the {ImageNet} dataset starts with a $7 \times 7$ px filter, followed by  $3 \times 3$ px and $1 \times 1$ px convolutions where the feature maps are subsampled 6 times. 
The same network on the {CIFAR-10} dataset exclusively uses $3 \times 3$ px convolutions and the feature maps are subsampled 2 times. 
Hard-coding the resolution hyper-parameters in the network for different datasets affects the extent of the receptive field, and the specific choices made can be restrictive.

In this paper we propose the N-JetNet which can replace CNN network design choices of filter sizes by learning these. 
We make use of scale-space theory~\cite{lindebergBook13scaleSpaceInCV}, where the resolution is modeled by the $\sigma$ parameter of the Gaussian function family and its derivatives. 
Gaussian derivatives allow a truncated Taylor series, called the N-Jet~\cite{florack1996gaussianLocalJet}, to model a convolutional filter~\cite{jacobsenCVPR16structuredRF} as a linear combination of Gaussian derivative filters, each weighted by an $\alpha_i$. 
We optimize these $\alpha$ weights instead of individual weights for each pixel in the filter, as done in a standard CNN. 
The choice of the basis cannot be avoided. 
In standard CNNs the choice is implicit: an N $\times$ N pixel-basis, whose size cannot be optimized, because it has no well-defined derivative to the error. 
In contrast, in the \model model the basis is a linear combination of Gaussian derivatives where the $\sigma$ parameter controls both the resolution and the filter size, and has a well-defined derivative to the effective filter and therefore to the error. 
This formulation allows the network to learn $\sigma$ and thus the network resolution. We exemplify our approach in \fig{fig1}.

To avoid confusions, we make the following naming conventions: throughout the paper we refer to `resolution' as the \emph{inner scale} as defined in \cite{koenderinkBioCyb84structOfIm}; `size' as the \emph{outer scale} \cite{koenderinkBioCyb84structOfIm} denoting the number of pixels of a filter or a feature map; and `scale' as the parameter controlling the resolution, which is the standard deviation $\sigma$ parameter of the Gaussian basis \cite{florack1996gaussianLocalJet}. 
The scale is different from the size of a filter: one can blur a filter and change its scale without necessarily changing its size.  
However, they are related as increasing the scale of an object (i.e. blurring) increases its size in the image (i.e. number of pixels it occupies).
Here we tie the filter size to the scale parameter by making it a function of $\sigma$.

We make the following contributions. 
(i) We exploit the multi-scale local jet for automatically learning the scale parameter, $\sigma$. 
(ii) We show both for classification and segmentation that our proposed \model model automatically learns the appropriate input resolution from the data. 
(iii) We demonstrate that our approach generalizes over network architectures and datasets without deteriorating accuracy for both classification and segmentation.

\section{Related work}
\noindent{\textbf{Multiples scales and sizes in the network.}}
Size plays an important role in CNNs. 
The highly successful inception architecture~\cite{SzegedyCVPR15inceptionGoingDeeper} uses two filter sizes per layer. 
Multiple input sizes can be weighted per layer~\cite{chenCVPR16attentionToScale}, integrated at the feature map level~\cite{ghiasiECCV16laplacian}, processed at the same time~\cite{bai2020multiscale,kanazawaNIPSW14locallyScaleInv,keCVPR17multigrid,li2019data} or even made to compete with each other~\cite{liaoPR17competeFilterSize}.
To process multiple featuremap sizes, spatial pyramids are used \cite{agustsson2020scale,lin2017feature,liu2020ipg,wang2020scale}, alternatively the best input size and network resolution can be selected over a validation set \cite{yangmutualnet}.  
Scale-equivarinat CNNs can be obtained by applying each filter at multiple sizes \cite{marcos2018scale}, or by approximating filters with Gaussian basis combinations \cite{sosnovik2019scale} where the set of scale parameters is not learned, but fixed. 
Unlike these works, we do not explicitly process our feature maps over a set of predefined fixed sizes. We learn a single scale parameter per layer from the data. 

Downsampling and upsampling can be modeled as a bijective function \cite{xiao2020invertible}, or made adaptive using reinforcement learning \cite{uzkent2020learning} and contextual information at the object boundaries \cite{marin2019efficient}.
The optimal size for processing an input giving the maximum classification confidence can be selected among multiple sizes \cite{li2012scale,li2011supervised}, or learned by mimicking the human visual focus \cite{sigurdsson2020beyond}, or minimizing the entropy over multiple input sizes at inference time \cite{wang2019dynamic}. 
Network architecture search can also be used for learning the resolution, at the cost of increased computations \cite{liu2019auto}.
Alternatively, the scale distribution can be adapted per image using dynamic gates \cite{li2020learning}, or by using self-attentive memory gates \cite{qi2016hierarchically}. 
The atrous~\cite{chenPAMI17deeplab,PapandreouCVPR15atrous} or dilated convolutions~\cite{YuICLR16multiScaleDilated,YuCVPR17dilatedResNet} design fixed versions of larger receptive fields without subsampling the image. 
These are extended to adaptive dilation factors learned through a sub-network \cite{zhang2020ascnet}.
Rather than only learning the filter size, we learn both the filter shape and the size jointly, by relying on scale-space theory.  

\smallskip\noindent{\textbf{Architectures accommodating subsampling.}}
A pooling operation groups features together before subsampling. 
Popular forms of grouping are average pooling~\cite{fukushima1986neural}, and max pooling~\cite{serreCVPR05hmax}. 
Average pooling tends to perform worse than max pooling~\cite{boureauCVPR10midLevelFeat,schererICANN10evaluationPooling} which is outperformed by their combination~\cite{boureauICML10theoreticalAnalysisPooling,leePAMI17generalizingPooling}.
Other forms include pooling based on ranking~\cite{shiNN16rankPool}, spatial pyramid pooling~\cite{heECCV14spatialPyrPool}, spectral pooling~\cite{rippelNIPS15spectralPool} and stochastic pooling~\cite{zeilerICLR13stochasticPool} and stochastic subsampling~\cite{zhaiCVPR17s3pool}. 
The recent BlurPool \cite{zhang2019making} avoids aliasing effects when sampling, while fractional pooling~\cite{grahamArxiv14fractionPool} subsamples with a factor of $\sqrt{2}$ instead of 2 which allows larger feature maps to be used in more network layers.  
All these pooling methods use hard-coded feature map subsampling. 
Our work differs, as we do not use fixed subsampling or strided convolution: we learn the resolution. 

\smallskip\noindent{\textbf{Fixed basis approximations.}}
Resolution in images is aptly modeled by scale-space theory~\cite{iijima1959basic,koenderinkBioCyb84structOfIm,WitkinIJCAI83scaleSpace}.  
This is achieved by convolving the image with filters of increasing scale, removing finer details at higher scales. 
Convolving with a Gaussian filter has the property of not introducing any artifacts~\cite{babaud1986uniqueness,duits2004axioms} and the differential structure of images can be probed with Gaussian derivative filters~\cite{florackIVC92scaleAndDiffStruct,lindebergBook13scaleSpaceInCV} which form the N-Jet~\cite{florack1996gaussianLocalJet}: a complete and stable basis to locally approximate any realistic image. 
Scale-spaces model images at different resolutions by a continuous one-parameter family of smoothed images, parametrized by the value of $\sigma$ of the Gaussian filter~\cite{koenderinkBioCyb84structOfIm}. In this paper we build on scale-space theory and exploit the differential structure of images to optimize $\sigma$ and thus learn the resolution. 

Various mathematical multi-scale image modeling tools have been used in convolutional networks. 
The classical work of Simoncelli \etal~\cite{simoncelli1992shiftable} proposes the steerable pyramid, defining a set of wavelets for orientation and scale invariance.
Similarly, the seminal Scattering transform~\cite{brunaPAMI13scatter,mallat2012groupInvScat} and its extensions~\cite{cotterMLSP17visualizing,sifreCVPR13rotScat} are based on carefully designed complex wavelet basis filters~\cite{mallat1999wavelet} with pre-defined rotations and scales giving excellent results on uniform datasets such as MNIST and textures. 
Using the Scattering transform as initialization for the first few layers of a CNN has recently~\cite{oyallonICCV17scalingScattering,singhCVPRw17efficientScatter}  been shown to also lead to good results on more varied datasets. 
Filters can also be approximated as a liner combination over a set of learned low-rank filter basis \cite{li2019learning}.
Recent work also starts with a filter basis and use a CNN to learn the filter weights. 
Examples include a PCA basis~\cite{ghiasiECCV16laplacian}, circular harmonics~\cite{WorrallCVPR17harmonicNets}, Gabors~\cite{luan2017gaborCNNs}, and Gaussian derivatives~\cite{jacobsenCVPR16structuredRF}.
In this paper we build on the Gaussian derivative basis~\cite{jacobsenCVPR16structuredRF} because it directly offers the tools of Gaussian scale-space to learn CNN resolution.

\begin{figure*}[t!]
    \includegraphics[width=1\linewidth]{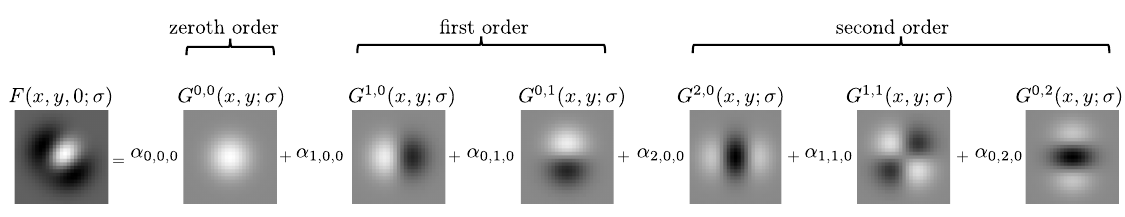}
    \caption{Representing local image structure with a linear combination of Gaussian basis filters. 
    The patch \textit{(left)}, $F(x,y,0; \sigma)$, is modeled by up to second order Gaussian derivatives, using six $\alpha$-coefficients.}
    \label{fig:gaussBasis}
\end{figure*}
\smallskip\noindent{\textbf{Learning kernel shape.}}
Current methods investigate inherent properties of CNN filters. 
Filters that go beyond convolution include non-linear Volterra kernels~\cite{ZoumpourlisICCV17nonLinConvFilt}, a learned image adaptive bilateral filter~\cite{jampaniCVPR16learnSparseHighDimFilt} and learned image processing operations~\cite{chenCVPR17fastIPfcn}. 
For convolutional CNN filters, Sun~\etal~\cite{sunECCV16designKernels} proposes an asymmetric kernel shape, which simulates hexagonal lattices leading to improved results. 
The active convolution by Jeon and Kim~\cite{jeonCVPR17activeConvLearnConvShape} and the deformable CNNs by Dai~\etal~\cite{daiICCV17deformable} offer an elegant approach to learn a spatial offset for each filter coefficient leading to flexible filters and improved accuracy. 
\cite{shocher2020discrete} learns continuous filters as functions over sub-pixel coordinates, allowing learnable resizing of the feature maps.
The hierarchical auto-zoom net~\cite{xiaECCV16autoZoomNet}, the scale proposal network~\cite{haoCVPR17scaleAwareFaceDet}, and the recurrent scale approximation network~\cite{liuICCV17recurScaleAprox} explicitly predict the object sizes and adapt the input size accordingly. 
Our work differs from all these methods because we learn both the filter shape and the size.

Most similar to us, \cite{shelhamer2019blurring,xiong2020variational} combine free-form filters with learned Gaussian kernels that can adapt the receptive field size. 
The recent work of Lindeberg~\etal \cite{lindeberg2020scalecovariant} uses Gaussian derivatives for scale-invariance, however the scales are fixed according to a geometric distribution.  
Dissimilar to these works we approximate the complete filter using a combination of Gaussian derivatives, while adapting the receptive field size.

\section{Learning network resolution }
\begin{figure*}[t]
\centering
\begin{tabular}{p{0.05\textwidth}p{0.9\textwidth}}
Original: & \includegraphics[width=1\linewidth]{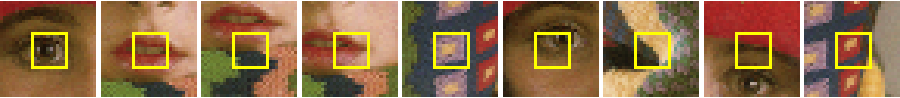} \\
Approx: & \includegraphics[width=1\linewidth]{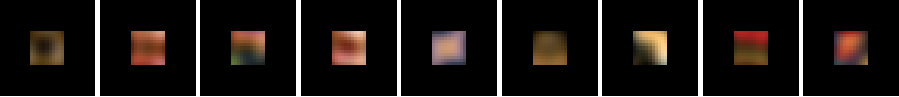} \\ 
\end{tabular}
\caption{ Illustration that a Gaussian basis can approximate local image structure. \textit{Top row:} The original cropped $11 \times 11$ patch. \textit{Bottom row:}  The approximation by a least-squares fit of the $\alpha$-coefficients using a third order RGB Gaussian basis with $\sigma=5$. The black border pixels are not evaluated in the least-squares fit. The approximation captures well a slightly blurred version ($\sigma=5$) of the original.}
\label{fig:trui11}
\end{figure*}

\begin{figure}[t]
\scriptsize
\centering
    \begin{tabular}{cc}
        \hspace{-10px}
      \includegraphics[width=0.25\textwidth]{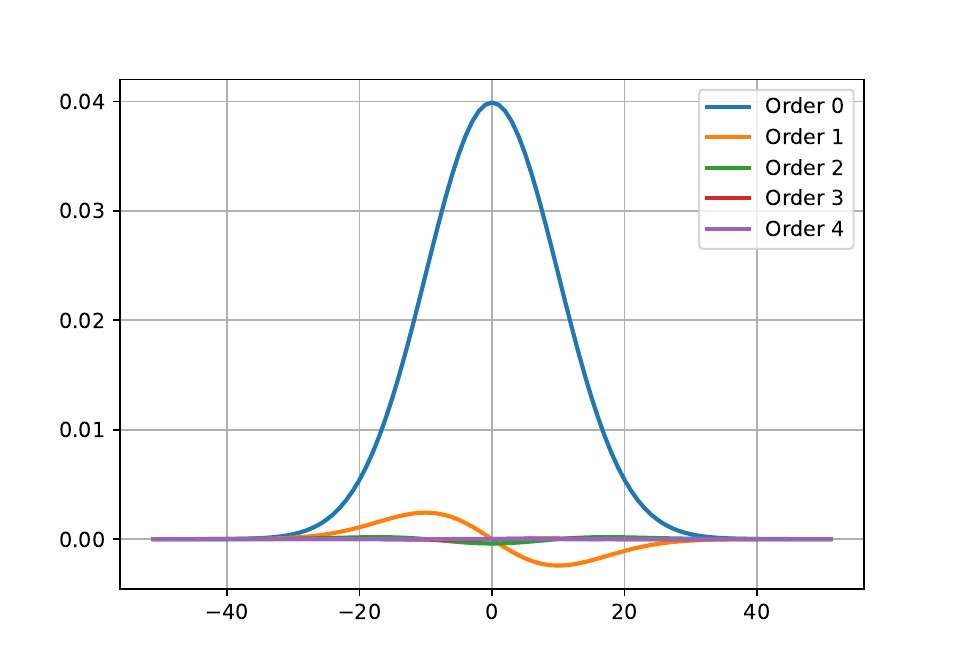} &
        \hspace{-15px}
      \includegraphics[width=0.25\textwidth]{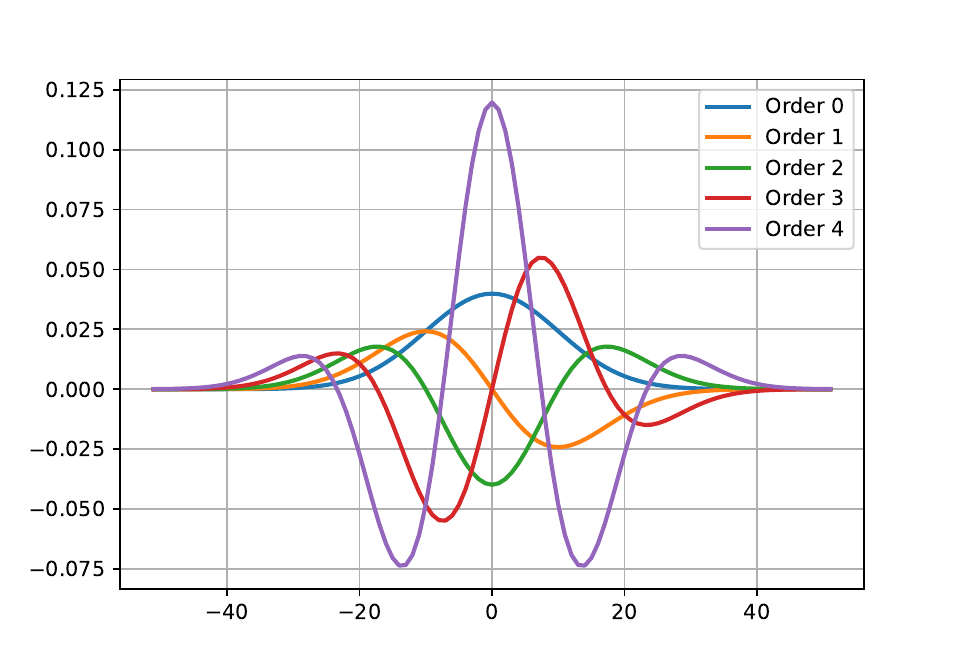} \\
      (a) Unnormalized & (b) Normalized \\
    \end{tabular}
  \caption{
    Effect of filter normalization. For unnormalized filters, the higher order filters are dwarfed by the lower order filters.
    Normalizing each basis filter of order i by multiplying with $\sigma^i$, ensures that the magnitude of each filter is approximately in the same range.
  }
  \label{fig:filter-normalization}
\end{figure}

\subsection{\textbf{Local image differentials at given scale}}
Scale-spaces~\cite{florackIVC92scaleAndDiffStruct,lindebergBook13scaleSpaceInCV,WitkinIJCAI83scaleSpace}
offer a general framework for modeling image structures at various scales. 
The resolution, or the \emph{inner scale}~\cite{koenderinkBioCyb84structOfIm} of an image is modeled by a convolution with a 2D Gaussian. The 1D Gaussian at scale $\sigma$ is given by $G(x; \sigma) =  \frac{1}{\sigma \sqrt{2 \pi}} e^{\frac{-x^2}{2\sigma^2}}$ which is readily extended to 2D as  $G(x, y; \sigma) = G(x;\ \sigma)\ G(y;\ \sigma)$. The local structure learned in deep networks~\cite{koenderinkBioCyb84structOfIm} is linked to the image derivatives. 
Image pixels are discretely measured, and do not directly offer derivatives. 
The linearity of the convolution operator allows~\cite{youngSV87gaussianDerivForSpatVis} to take an exact derivative of a slightly smoothed function $f$ with a Gaussian kernel $G(.;\sigma)$ with scale $\sigma$: 
\begin{align}
\frac{\partial (f(x) \ast G(x;\sigma) )}{\partial x}   = \frac{\partial G(x;\sigma)}{\partial x} \ast f(x),
\end{align}
where $\ast$ denotes a convolution. 
This allows taking image derivatives by convolving the image with Gaussian derivatives. 
Gaussian derivatives in 1D at order $m$ and scale $\sigma$ can be defined recursively using the Hermite polynomials~\cite{martens1990hermite}:
\begin{align}
G^m(x; \sigma) &= \frac{\partial^m G(x,\sigma)}{\partial x^m}  \\ \nonumber
    & = \left(\frac{-1}{\sigma \sqrt{2} }\right)^m H_m\left(\frac{x}{\sigma \sqrt{2}}\right) G(x; \sigma),  
\end{align}
where $G(x; \sigma)$ is the Gaussian function and $H_m(x)$ the $m$-th order Hermite polynomial, recursively defined as $H_i(x) = 2x H_{i-1}(x) - 2(i-1) H_{i-2}(x) ; \    H_0(x) = 1 ; \ H_1(x) = 2x$. 
We define 2D Gaussian derivatives by the product of the partial derivatives on $x$ and on $y$: 
\begin{align}
G^{i,j}(x, y; \sigma) &= \frac{\partial^{i+j}G(x, y;\ \sigma)}{\partial x^{i} \partial y^{j}}\ \\ \nonumber
&= \ \frac{\partial^{i}G(x;\ \sigma)}{\partial x^{i}}\ \frac{\partial^{j}G(y;\ \sigma)}{\partial y^{j}}.
  \label{eq:seperability}
\end{align}

\subsection{\textbf{Multi-scale local N-Jet for modeling local image structure}}
A discrete set of Gaussian derivatives up to $n^\text{th}$ order, $\{ G^{i,j}(x,y; \sigma) \mid 0 \leq i+j \leq n \}$, can be used in a truncated local Taylor expansion to represent the local scale-space near any given point with increasing accuracy~\cite{florack1996gaussianLocalJet}. 
This allows us to approximate a filter $F(x)$ around the point $a$ up to order $N$ as:
\begin{align}
F(x) = \sum^N_{i=0} \frac{ \frac{\partial^i}{\partial x^i} F(a)}{i!} (x - a)^i + \frac{ R(a)}{(N+1)!} (x - a)^{N+1}, 
\end{align}
where $R$ is the residual term that corresponds to the approximation error.  
By absorbing the polynomial coefficients into a value $\alpha$, we arrive at a linear combination of Gaussian derivative basis filters which can be used to approximate image filters, as illustrated in~\fig{gaussBasis}. For filter $F(x,y,c)$ at position $(x,y)$ and color channel $c$ the approximation is:
\begin{align}
  F(x, y, c; \sigma)\ &=\ \sum_{\substack{0\ \le\ i,\ 0\ \le\ j}}^{i + j\ \le\ N}\ \alpha_{i, j, c}\ \frac{\partial^{i+j}}{\partial x^{i} \partial y^{j}} G\left(x, y;\ \sigma \right) \nonumber \\
    & + R(x,y, c; \sigma),
\label{eq:taylor}
\end{align}
where $R$ is the residual error, ignored here. 
Optimizing the $\alpha$ parameters allows us to switch from learning pixel weights as commonly used in CNNs, to learning the weights of the Gaussian basis filters. 
We show some examples in \fig{trui11} where we optimize the $\alpha$ parameters of an order-3 RGB Gaussian derivative basis with $\sigma=5$ to least squares fit an $11 \times 11$ px patch. 
Results show that the fit can approximate well a slightly blurred version of the original patch. Because $\sigma=5$ we cannot recover a perfectly sharp faithful copy of the original patch.
\subsection{\textbf{Learning receptive field size}}
We have all ingredients to learn the resolution in a convolutional deep neural network (CNN). 
Resolution is bounded by the size of the CNN filters. 
We can now dynamically adapt the resolution during training.

\smallskip \noindent {\textbf{Scale-invariant basis normalization.}}
The filter responses of Gaussian derivatives decay with order, as depicted in figure~\ref{fig:filter-normalization}.(a).
Following \cite{florack1996gaussianLocalJet}, we make the Gaussian derivatives scale-independent by multiply each $i$-th order partial derivative by $\sigma^i$.
This brings the magnitude of basis filters in approximately the same range, as illustrated in \fig{filter-normalization}.(b). 

\smallskip \noindent \textbf{Learning scale and filter size.} 
The network resolution depends on the parameter $\sigma$, determining the inner scale of the Gaussian derivative basis. 
The chain-rule for differentiation allows to express the derivative of the 
error $J$ with respect to $\sigma$ as the product of two terms: 
$\frac{\partial J}{\partial \sigma} = \frac{\partial J}{\partial F} \cdot \frac{\partial F}{\partial \sigma}$. 
The first term is the derivative of the error with respect to the filter and it is found by error-backpropagation, as standardly done. 
The second term is the derivative of the filter with respect to $\sigma$ and can be found by differentiating \eq{taylor} with respect to $\sigma$. 
Similarly, the value of the Gaussian basis mixing coefficients, $\alpha_{i,j,c}$ can be found by differentiating the filter $F$ with respect to the coefficients $\alpha$.

In practice we cannot work with continuous filters. 
Therefore, we need to clip the filters to a finite size to perform the convolution. 
The size $s$ of the filter follows the formula: $ s = 2 \left \lceil\ k \sigma\ \right \rceil + 1$, where $k$ determines the extent of the local N-Jet approximation and is experimentally set.
By tying the filter size to the scale parameter, we only need to change $\sigma$ and adapt both the scale controlling the network resolution, and the size defining the spatial extent of the filters. 

\section{Experiments}

\subsection{\textbf{Exp (A): \model for Image Classification}} 
\label{sec:classification_results}

\noindent\textbf{Safely subsampling for image classification.} 
The receptive field size is also altered through subsampling, pooling, or strided convolution. 
For classification models we remove all subsampling operations in the network and add a \emph{\safespl} operation.
If the resolution is low (\ie, the $\sigma$ value is high) then there is no need to keep the feature map at full size, and it can safely be subsampled, to improve memory and speed. 
For a feature map of size $s$, we subsample the feature map to a new size $\bar{s}$, where we half its current size as a function of $\sigma$ as: $\bar{s} = s \left( \frac{1}{2} \right)^{\sigma/r}$,
where $r$ is the \emph{\safespl} hyper-parameter.
We apply \emph{\safespl} for all models except for the very deep networks: \emph{Resnet-110} and \emph{EfficientNet}, where it is reducing the feature map sizes too much.

\begin{figure*}[t]
	\small
	\centering 
    \begin{tabular}{ c @{\hskip 0.9in} c }
		{ \scriptsize
		\begin{tabular}{ c | c | c }
			  \includegraphics[height=0.5cm]{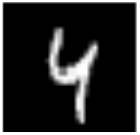} & 
			  \includegraphics[height=0.75cm]{mnist-4} & 
			  \includegraphics[height=1.0cm]{mnist-4} \\
			  \toprule
			  \textbf{1.0x MNIST} & \textbf{1.5x MNIST} & \textbf{2.0x MNIST} \\
			  \midrule
			  \multicolumn{3}{c}{structured conv, order 4, filters 16} \\
			  \midrule
			  \multicolumn{3}{c}{batch norm, relu} \\
			  \midrule
			  2x2 max pool & 3x3 max pool & 4x4 max pool \\
			  stride 2 & stride 3 & stride 4 \\      
			  \midrule
			  \multicolumn{3}{c}{fully-connected, softmax} \\
			  \bottomrule
		\end{tabular}} &
		\begin{tabular}{ c }
		\includegraphics[width=0.35\textwidth]{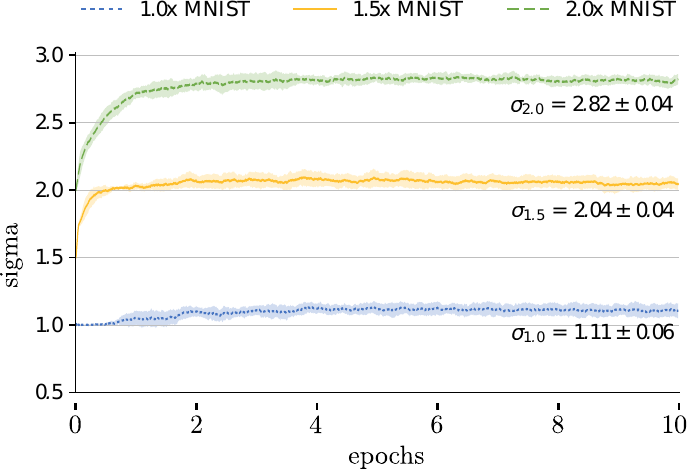}\\
		\end{tabular} 
        \\
		(a) Network architecture. & (b) Learned $\sigma$ on multi-size \emph{MNIST}.\\
	\end{tabular}
		\caption{ 
		\textbf{Exp 1.2(A):} 
		(a) Toy architecture used for testing whether we can learn the correct data resolution from the inputs. 
		(b) Estimated Gaussian basis scale, $\sigma$, on \emph{MNIST} resized $1\times$, $1.5\times$ and $2\times$.
		The estimated $\sigma$ follows the resizing of the data. 
	    }
	\label{fig:validation2}
\end{figure*}

\begin{figure}[t]
\centering
  \includegraphics[width=0.45\textwidth]{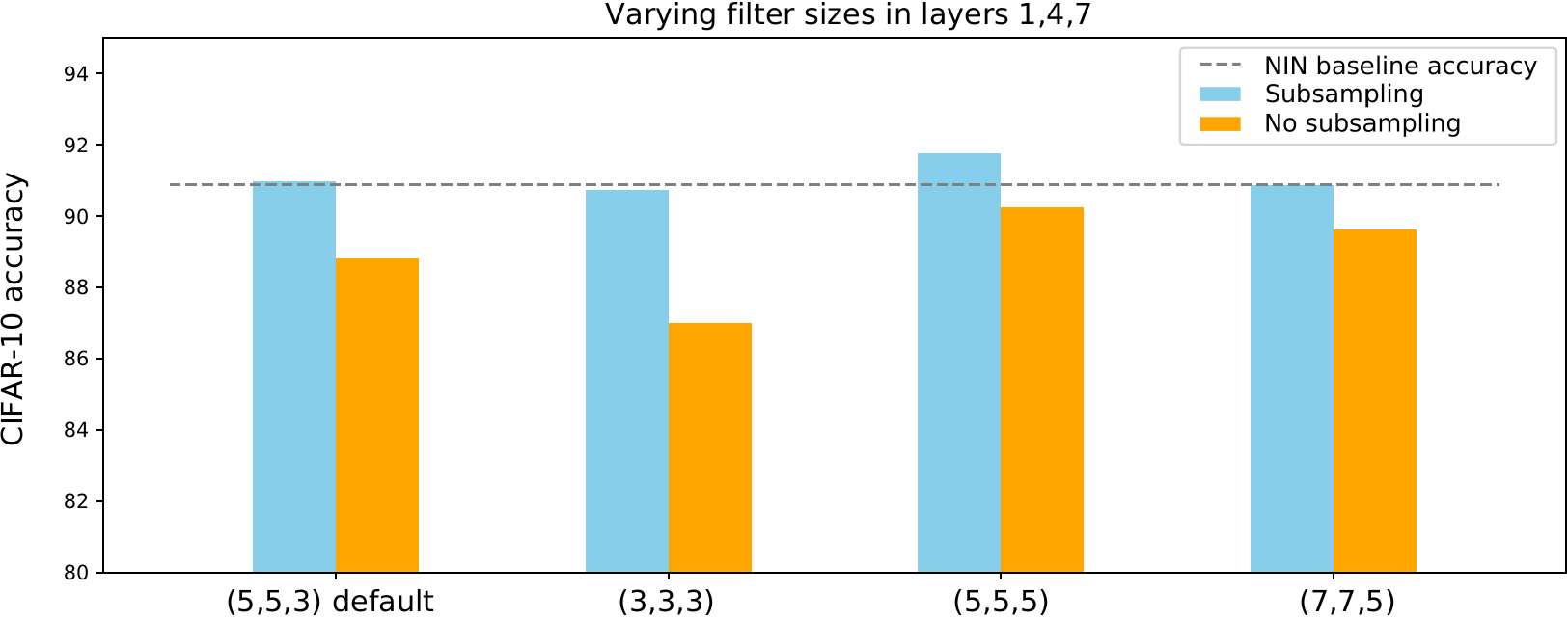}
  \caption{
		\textbf{Exp 1.1(A):} The impact on accuracy when varying filter sizes and feature map size for the \emph{NIN} baseline on \emph{CIFAR-10}. 
		Smaller filter sizes are more affected by the removal of the subsampling. 
		Setting the resolution hyper-parameters wrong can deteriorate the accuracy.
  }
  \label{fig:validation1}
\end{figure}

\smallskip\noindent \textbf{Experimental setup.}
We validate our approach on three standard datasets: \emph{CIFAR-10}, \emph{CIFAR-100} \cite{krizhevsky2009learning} and \emph{SVHN} \cite{netzer2011reading}. 
and multiple network architectures: \emph{NIN} (Network in Network) \cite{lin2013network}, \emph{ALLCNN} \cite{springenbergICLR15striveSimpleAllCNN} and \emph{Resnet-32}, \emph{Resnet-110} \cite{he2016deep}, as well as the recent \emph{EfficientNet} \cite{tan2019efficientnet}. 
To derive our \model models, we replace all the normal convolutional layers with variants of our \layer layers.
When using \safespl we remove all pooling layers, and set the stride to 1 in all network layers. 
In all our \model models we set the order of the Taylor series approximation to $3$, unless otherwise specified.
For all the models reported, we add a batch normalization layer after the convolutional layers, for robustness, and use the momentum SGD with the momentum set to $0.9$.
We train for the number of epoch reported in the literature. 
For the \emph{NIN} baseline model we found the best starting learning rate to be $0.5$, while for the \emph{ALLCNN} baseline $0.25$.
We use the same starting learning rates in our \model models.
For our \emph{\model-NIN} model we use an $L_2$ regularization weight over the Gaussian mixing coefficients, $\mathbb{\alpha}$, set to $0.01$, 
while for \emph{\model-ALLCNN} we regularize the $\mathbb{\alpha}$-s with a weight decay of $0.001$. 
When training our \emph{\model-Resnet} models we use an $L_2$ regularization weight over the Gaussian mixing coefficients, $\mathbb{\alpha}$, set to $0.0001$, and a starting learning rate of $0.1$ as indicated in \cite{he2016deep}.
We evaluate on relatively small datasets, and therefore we use the lightweight version of \emph{Resnet} where the first block has 16 channels and the last block 64, while for the deeper \emph{Resnet-110} models we use bottleneck blocks with a 4$\times$ channel expansion.
For both the baseline \emph{EfficientNet} and our \emph{\model-EfficientNet} we train the models from scratch, and given the small datasets we use the smallest model B0 \cite{tan2019efficientnet}.
For the \emph{EfficientNet} baseline we use a $0.01$ learning rate and a batch size of 32 and we rescale the inputs to $224\times 224$ px, since otherwise the model performs poorly, maybe due to the large subsampling.
In our \emph{\model-EfficientNet} we keep the input images to their original size. 
For our deeper models \emph{\model-EfficientNet} and \emph{\model-Resnet-110} we use batches of 16 and a learning rate of $0.001$.
\footnote{We will provide the \model code\break at http://github.com/SilviaLauraPintea/N-JetNet}.

\medskip\noindent\textbf{Experiment 1(A): Validation}

\noindent \textbf{Experiment 1.1(A): Do resolution hyper-parameters really matter?} 
We test our assumption that filter sizes and feature map sizes affect accuracy.
For this we use the \emph{NIN} baseline trained on \emph{CIFAR-10}. 
We vary the filter sizes in the layers of the \emph{NIN} which are not $1\times 1$ convolutions, and we reset the strides to 1 in all layers, to remove the feature map subsampling.
\fig{validation1} shows the impact of changing the filter sizes and removing the subsampling.
The smaller filter sizes, as in the case when all filter sizes are set to 3, are affected to a greater degree by the removal of the subsampling because they have a smaller receptive field.    
Selecting the correct filter sizes impacts the overall classification accuracy, and an exhaustive search over all possible filter size combinations is not feasible.
This validates the need for learning filter sizes. 

\smallskip\noindent \textbf{Experiment 1.2(A): Can the image resolution be learned?}
To test resolution learning, we create a toy network architecture depicted in \fig{validation2}(a).
We train the toy architecture on \emph{MNIST} when resizing the images $1\times$, $1.5\times$, and $2\times$. 
\fig{validation2}.(b) shows the learned Gaussian basis scale, $\sigma$, per setup. 
The $\sigma$ values learned for the images resized by $1.5$ and $2$ do not directly correspond to these values because the operations of sampling and resizing are not commutative: we first discretized the continuous signal into an image and subsequently subsampled it.  
However, the relative ratio between the learned scales is close: $(2.0/1.5)  \sigma_{1.5} = 2.81 \pm 0.04 \approx \sigma_{2.0} = 2.82 \pm 0.04$. 
The learned $\sigma$ values follow the input resizing, thus the correct filter scales and sizes can be learned from the input. 

\medskip\noindent\textbf{Experiment 2(A): Model choices}

\smallskip\noindent\textbf{Experiment 2.1(A): Learning sigma.}
We test the effect of $\sigma$ on the performance on the \emph{CIFAR-10} dataset, using the \emph{NIN} backbone.
We fix the spatial extent, $k$, to $2$ and vary sigma in the set $\{0.5, 1.0, 2.0\}$.
\tab{intern1} shows that a wrong setting of $\sigma$ can influence the classification accuracy up to 3\%.
The \safespl setting is affected more by the choice of $\sigma$ than the baseline subsampling as it relies on the value of $\sigma$ when deciding how much to subsample the input feature maps. 
Overall, we note that $\sigma=1.0$ achieves the best performance on this setting, therefore we use this value when initializing $\sigma$ during the learning in our \model models.
\begin{table}[t]
    \caption{
        \textbf{Exp 2.1(A):} The effect on \emph{CIFAR-10} of varying the filter scale, $\sigma$.
        Having a sub-optimal filter scale $\sigma$ can decrease accuracy up to 3\%. 
        The \safespl is slightly more sensitive to $\sigma$ than the baseline sampling.
	}
\centering
	\begin{tabular}{l @{\hskip 0.2in} c @{\hskip 0.2in} c @{\hskip 0.2in} c}
			\toprule
									&	\multicolumn{3}{c}{$\sigma$} \\ \cmidrule(l){2-4}
			Sampling			& $\sigma = 0.5$	& $\sigma = 1.0$	&  $\sigma = 2.0$ \\ \midrule
			Baseline			& 88.76\%			& 90.25\%	        & 87.39\% \\
			\Safespl			& 86.29\%           & 89.50\%	        & 87.26\% \\
			\bottomrule
	\end{tabular} 
	\label{tab:intern1} 
\end{table}

\smallskip\noindent\textbf{Experiment 2.2(A): \Safespl.}
We test the importance of the hyper-parameter $r$ in the \safespl, with respect to the classification accuracy on \emph{CIFAR-10} using a \emph{NIN} backbone. 
For this experiment we learn the filter scale $\sigma$ and set $k=2$.
We fix the hyper-parameter $r$ to one of the values in the set $\{2.0, 4.0, 6.0\}$.
\tab{intern3} shows the effect on accuracy of different settings of $r$.
We also show the runtime needed to train the network for different $r$ settings. 
As the value of $r$ increases the accuracy also increases, however also the feature map sizes in the layers of the network increase, which affect the overall computational time. 
For our subsequent experiments we select $r=4.0$ as a trade-off between accuracy and training speed.
\begin{table}[t]
	\caption{
	\textbf{Exp 2.2(A):} The importance of the hyper-parameter $r$ of the \safespl on the \emph{CIFAR-10} accuracy.
	The accuracy slightly increases as the hyper-parameter $r$ of the \safespl increases. 
	}
    \centering
    \resizebox{1.0\linewidth}{!}{
	\begin{tabular}{l @{\hskip 0.30in} c @{\hskip 0.2in} c @{\hskip 0.2in} c}
			\toprule
					 &			\multicolumn{3}{c}{\Safespl hyper-parameter}\\ \cmidrule(l){2-4}
			&  $r = 2.0$	& $r = 4.0$		  &  $r = 6.0$ \\ \midrule
			Accuracy & 91.59\%	& 91.60\%         & 91.63\% \\ 
			\multicolumn{1}{l}{Training time} & $\approx$129.91	min. &  $\approx$179.43 min.  & $\approx$193.48 min. \\
			\bottomrule
	\end{tabular} 
    }
	\label{tab:intern3} 
\end{table}

\medskip\medskip\noindent\textbf{Experiment 3(A): Generalization ability}

\smallskip\noindent\textbf{Experiment 3.1(A): Generalization to other datasets.} 
\begin{table}[t]
    \centering
	\caption{ 
		\textbf{Exp 3.1(A):} Dataset generalization.
		Comparison of our \emph{\model-NIN} and baseline \emph{NIN} on the \emph{CIFAR-100} and \emph{SVHN} datasets.
        For the baseline model we only report the best performance we obtain rerunning the models, while for our \model models we report mean and standard deviations over 3 runs.
        We achieve comparable classification accuracy with the baseline.
	}
    	\begin{tabular}{l @{\hskip 0.2in} l @{\hskip 0.2in} l @{\hskip 0.2in}}
			\toprule
			                & \emph{NIN} \cite{lin2013network} & \emph{\model-NIN} (Ours)\\ \midrule
			\emph{SVHN}          & 98.17\%                   & 97.55\% ($\pm$0.08) \\
			\emph{CIFAR-10}      & 90.89\%                   & 91.60\% ($\pm$.08) \\
			\emph{CIFAR-100}     & 66.14\%                   & 68.42\% ($\pm$0.31) \\
			\bottomrule
		\end{tabular} 
		\label{tab:dataset} 
\end{table}
\begin{figure}[t]
	\includegraphics[width=0.45\textwidth]{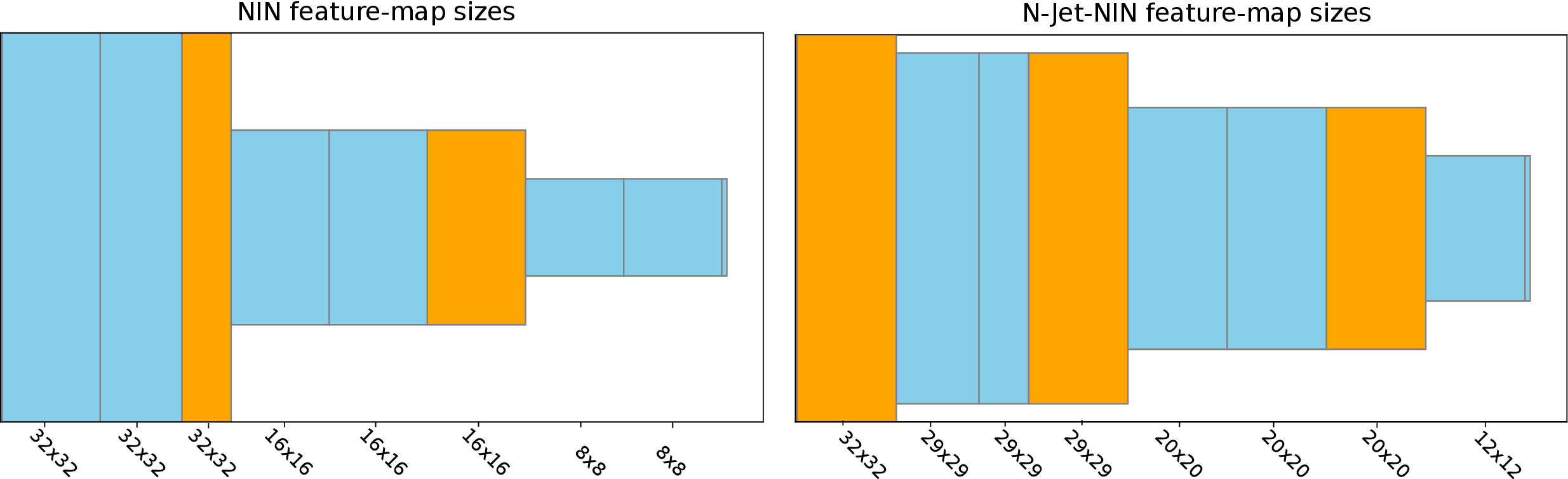}\\
	\caption{
	    \textbf{Exp 3.1(A):} Dataset generalization.
		The baseline feature map sizes compared to the learned feature maps sizes by our \emph{\model-NIN} on \emph{CIFAR-10}.
		We show in orange the subsampling layers. 
		Safe-subsampling dynamically finds the appropriate feature map size.	
	}
	\label{fig:dataset} 
\end{figure}
We compare our \emph{\model-NIN} method with the baseline \emph{NIN} \cite{lin2013network}.
We test the generalization properties of our method by also reporting scores on two other datasets: \emph{CIFAR-100} and \emph{SVHN}.
\tab{dataset} shows the classification results of our \emph{\model-NIN} when compared with the baseline \emph{NIN}. 
We report mean and standard deviations over $3$ runs for our method. 
We show in \fig{dataset} the hard-coded sizes of the baseline feature maps, versus the sizes learned by our \emph{\model-NIN} on \emph{CIFAR-10}. 
The performance of \emph{\model-NIN} is comparable with the baseline performance, while dynamically learning the appropriate feature map size. 

\smallskip\noindent\textbf{Experiment 3.2(A): Generalization to other models.} 
\begin{table}[t]
	\caption{ 
		\textbf{Exp 3.2(A):} Architecture generalization.
		The classification accuracy on \emph{CIFAR-10}, \emph{CIFAR-100} when comparing the baseline models with our proposed \emph{\model-ALLCNN}, \emph{\model-Resnet-32}, and two deeper models: \emph{\model-Resnet-110} and \emph{\model-EfficientNet}.
        For the baseline models we report the best performance, while for our \model models we report mean and standard deviations over 3 runs. 
       We also show the number of parameters for each model.
       \emph{\model} nets obtain comparable accuracy to the baselines while reducing the number of parameters for order 2.
    }
    \centering
    \resizebox{1.0\linewidth}{!}
    {
    {
    \begin{tabular}{llll}
			\toprule
            & {ALLCNN \cite{springenbergICLR15striveSimpleAllCNN}} & \multicolumn{2}{c}{\model-ALLCNN (Ours)}\\
            & & Order 3 & Order 2 \\ \cmidrule(l){2-2}\cmidrule(l){3-4}
    		\# params & 0.97 M & 1.07 M & 0.66 M\\
			\emph{CIFAR-10}  &  91.87\% & 92.48\% ($\pm 0.134$) & 89.91\% ($\pm 0.032$) \\  
			\emph{CIFAR-100} &  67.24\% & 67.62\% ($\pm 0.863$) & 65.17\% ($\pm 0.228$) \\ \midrule
			
			& Resnet-32 \cite{he2016deep} & \multicolumn{2}{c}{\model-Resnet-32 (Ours)} \\ 
			& & Order  3 & Order 2 \\ \cmidrule(l){2-2}\cmidrule(l){3-4}
			\# params &  0.47 M & 0.52 M & 0.31 M \\
			\emph{CIFAR-10}  & 92.30\% & 92.28\% ($\pm 0.260$) & 89.49\% ($\pm 0.304$) \\
			\emph{CIFAR-100} & 67.89\% & 67.59\% ($\pm 0.278$) & 65.14\% ($\pm 0.619$) \\ \midrule

			& Resnet-110 \cite{he2016deep} & \multicolumn{2}{c}{\model-Resnet-110 (Ours)} \\ 
			& & Order  3 & Order 2 \\ \cmidrule(l){2-2}\cmidrule(l){3-4}
    		\# params & 6.90 M & M 7.29 & 5.74 M \\
			\emph{CIFAR-10}  & 92.83\% & 93.71\% ($\pm 0.337$) & 93.52\% ($\pm 0.043$)\\
			\emph{CIFAR-100} & 73.53\% & 71.73\% ($\pm 0.203$) & 73.66\% ($\pm 0.295$) \\ \midrule

    		& EfficientNet \cite{tan2019efficientnet} & \multicolumn{2}{c}{\model-EfficientNet (Ours)} \\
    		& & Order  3 & Order 2 \\ \cmidrule(l){2-2}\cmidrule(l){3-4}
			\# params &  3.60 M &  3.51 M  & 3.48 M \\
			\emph{CIFAR-10} & 92.64\% & 93.51\% ($\pm 0.110$) & 93.71\% ($\pm 0.029$) \\
			\emph{CIFAR-100} & 76.19\% & 75.22\% ($\pm 0.163$) & 76.17\% ($\pm 0.409$) \\
			\bottomrule
    \end{tabular}
    }}
	\label{tab:network} 
\end{table}
\begin{figure}[t]
        \centering
		\includegraphics[width=0.45\textwidth]{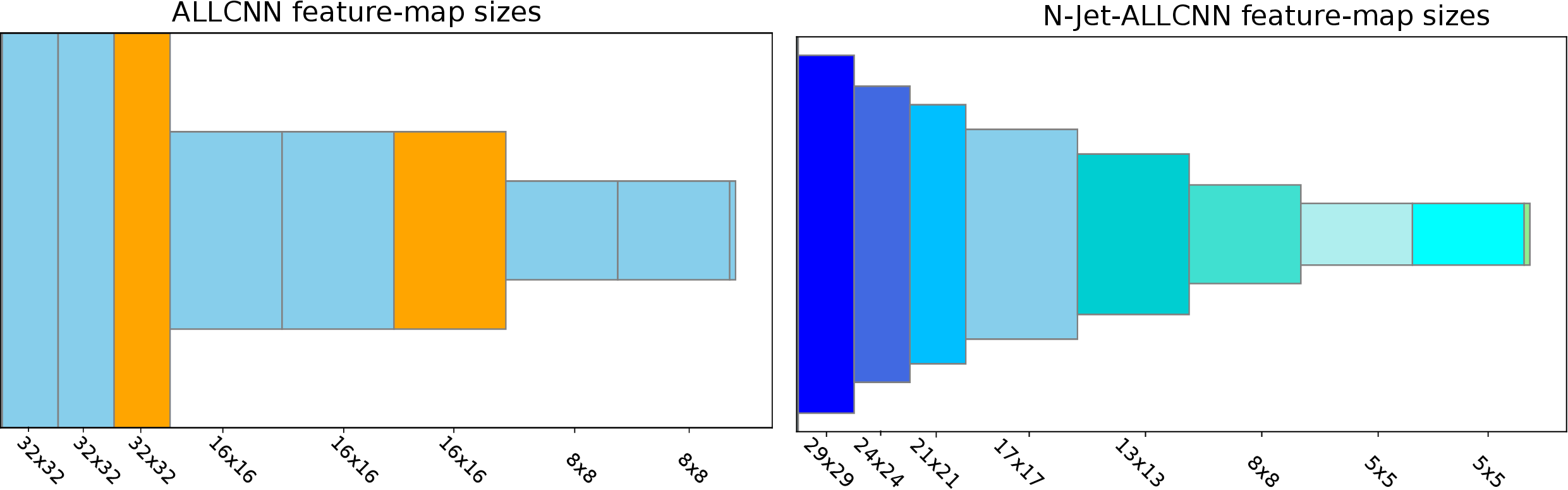}
		\caption{
		\textbf{Exp 3.2(A):} Architecture generalization.
		The baseline feature map sizes when compared to the learned \model feature map sizes. 
		We show in different colors the layers at which the size changes. 
        For the standard architecture the subsampling layers are orange.  
		We can find at every layer the appropriate subsampling level.
		\label{fig:network} 
		}
\end{figure}
\footnotetext{We use torchinfo (https://github.com/tyleryep/torchinfo) to enumerate all the parameters.}
To test the generalization of our \layer layer to different network architectures, we use the \emph{ALLCNN} \cite{springenbergICLR15striveSimpleAllCNN}, Resnet \cite{he2016deep}, and the recent \emph{EfficientNet} \cite{tan2019efficientnet} backbone network architectures.
For \emph{\model-ALLCNN} we use \safespl at every layer, while for \emph{\model-Resnet-32} only at the layers where the original network subsamples. 
\tab{network} shows the classification accuracy of the baseline models tested by us on \emph{CIFAR-10}, \emph{CIFAR-100}, when compared with our \emph{\model} models.
We report mean and standard deviation over $3$ repetitions for our models, as well as the number of parameters.
Using our proposed \model layers gives similar accuracy to the standard convolutional layers, while avoiding the need to hard-code the filter sizes.
For an approximation of order 3 in the \model, there is a small increase in the number of parameters compared to the baseline, except for the \emph{EfficientNet} which uses also kernel sizes larger than $3\times3$ px.
When employing larger models -- \emph{\model-Resnet-110} and \emph{\model-EfficientNet} -- a Gaussian basis combination of order 2 is sufficient to obtain an accuracy comparable to the baseline models, while reducing the number of parameters.  
In \fig{network} we show the baseline \emph{ALLCNN} feature map sizes when compared to the \emph{\model-ALLCNN} learned feature map sizes on \emph{CIFAR-10}.
The \model model has similar classification accuracy when compared to the \emph{ALLCNN} baseline, while learning at every layer the befitting feature map size. 
Applied at every layer, the \safespl makes the subsampling continuous and smooth, compared to the baseline.   
 

\begin{table}[t]
\caption{
    \textbf{Exp 4(A):} Comparison with scale-invariant methods. We evaluate on MNIST and MNIST resized 4$\times$ our \model, standard convolutions of varying filter sizes, as well as using Atrous convolutions \cite{chenPAMI17deeplab} and deformable convolutions \cite{chenPAMI17deeplab}.
    Our \model performs well on MNIST $\times$ 4 despite the increase in size.}
\centering
\resizebox{1\linewidth}{!}{
\begin{tabular}{l@{\hskip 0.3in} l@{\hskip 0.3in} l  }
\toprule
                                        & \multicolumn{2}{c}{2-layer architecture}\\  
CNN                                     & MNIST                 & 4 $\times$ MNIST\\\cmidrule(l){2-3}
Standard $3 \times 3$                   & 97.04\% ($\pm$ 0.22)  & 86.27\% ($\pm$ 3.44)\\
Standard $5 \times 5$                   & 98.62\% ($\pm$ 0.08)  & 88.67\% ($\pm$ 5.97)\\                
Standard $9 \times 9$                   & 98.93\% ($\pm$ 0.08)  & 95.84\% ($\pm$ 0.76)\\
Standard $11 \times 11$                 & 98.72\% ($\pm$ 0.20)  & 95.75\% ($\pm$ 1.50)\\   \midrule                   
Atrous \cite{chenPAMI17deeplab}         & 98.47\% ($\pm$ 0.25)  & 89.87\% ($\pm$ 3.67)\\               
Deformable \cite{daiICCV17deformable}   & 97.54\% ($\pm$ 0.46)  & 84.09\% ($\pm$ 0.84)\\               
\model (Ours)                                  & 99.05\% ($\pm$ 0.04)  & 98.11\% ($\pm$ 0.16)\\
\end{tabular}
}
\centering
\resizebox{1\linewidth}{!}{
\begin{tabular}{l@{\hskip 0.3in} l@{\hskip 0.3in} l}
\toprule
                                        & \multicolumn{2}{c}{4-layer architecture}\\  
CNN                                     & MNIST                 & 4 $\times$ MNIST \\ \cmidrule(l){2-3}
Standard $3 \times 3$                   & 98.49\% ($\pm$ 0.20)  & 86.53\% ($\pm$ 7.15) \\
Standard $5 \times 5$                   & 98.54\% ($\pm$ 0.37)  & 97.47\% ($\pm$ 0.41) \\                
Standard $9 \times 9$                   & 98.91\% ($\pm$ 0.18)  & 94.23\% ($\pm$ 4.01) \\
Standard $11 \times 11$                 & 98.81\% ($\pm$ 0.20)  & 96.21\% ($\pm$ 1.40) \\   \midrule                   
Atrous \cite{chenPAMI17deeplab}         & 98.40\% ($\pm$ 0.38)  & 91.28\% ($\pm$ 1.88) \\               
Deformable \cite{daiICCV17deformable}   & 98.91\% ($\pm$ 0.27)  & 86.45\% ($\pm$ 1.32) \\               
\model (Ours)                                 & 99.37\% ($\pm$ 0.05)  & 98.87\% ($\pm$ 0.21) \\
\bottomrule
\end{tabular}
}
\label{tab:scale_table}
\end{table}
\medskip\noindent\textbf{Experiment 4(A): Comparison to scale-invariant methods.}
\noindent We evaluate on the normal sized $28 \times 28$~px MNIST and on MNIST resized by a factor of 4 with a size of $112 \times 112$~px. We compare against a standard CNN with varying filter sizes, and against the Deformable CNN~\cite{daiICCV17deformable}, as well as Atrous (dilated) convolutions~\cite{chenPAMI17deeplab}. 
We consider 2 and 4-layer toy architectures containing only convolutional layers followed by ReLU activations. 
Results in table~\ref{tab:scale_table} show that the standard CNN performs well on MNIST, yet results are sensitive to the filter size for 4 $\times$ MNIST. 
The Deformable CNN \cite{daiICCV17deformable} is also affected by the change in image size.
Our intuition is that the Deformable CNN still relies on the initial $3 \times 3$ convolutions and optimizing the offsets is difficult under large size changes in the input. 
For Atrous CNN the dilation factor has to be hard-coded,  and we use a dilation factor of 2, as using 4 would imply including prior knowledge. 
The Atrous performance is also affected by the change in input size. In contrast, our N-Jet model is able to learn the correct resolution and is more accurate.

\subsection{\textbf{Exp (B): \model for Image Segmentation}}
\noindent\textbf{Learning the receptive field size for segmentation.} 
Multi-scale information processing is heavily used in modern segmentation architectures, and seems to be an important performance booster~\cite{chenCVPR16attentionToScale}. 
Here, we focus on two popular mechanisms for multi-scale processing, namely the merging of information at different scales via skip connections in \emph{U-Net} architectures~\cite{UNet}, and the pooling of information at different scales via atrous spatial pyramid pooling (ASPP) layers in \emph{DeepLab} architectures \cite{chenPAMI17deeplab,Deeplabv3}.

Similar to the classification experiments (\sect{classification_results}), we replace the fixed-size convolutional filters of baseline networks with the \model~definition, where we learn the size and scale of the filters in the convolution operations during training.

\medskip\noindent\textbf{Experiment 1(B). Segmentation with U-Net}

\begin{figure}[t]
\scriptsize
\centering
    \begin{tabular}{cc}
      \includegraphics[trim=80 0 80 0,clip, width=0.2\textwidth]{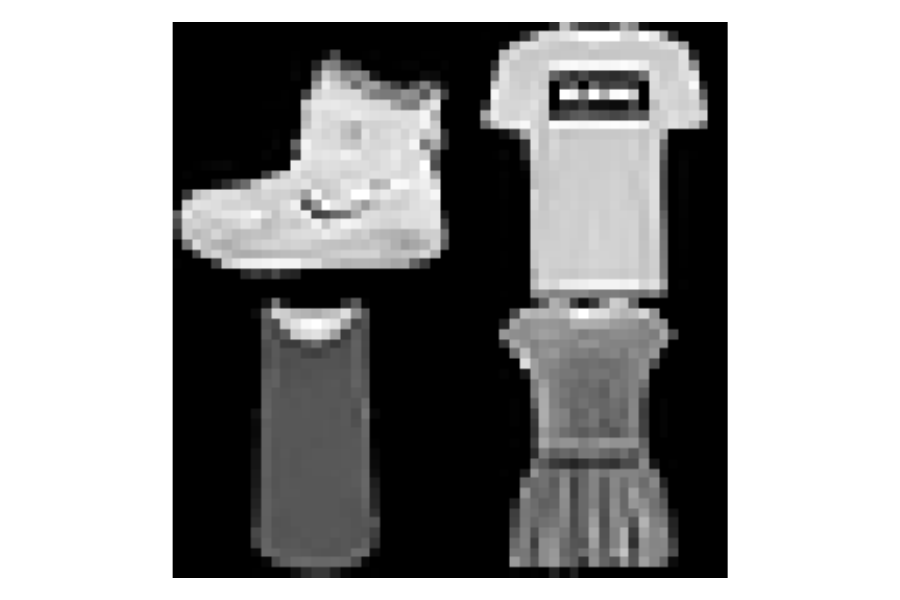} & 
      \includegraphics[trim=80 0 80 0,clip, width=0.2\textwidth]{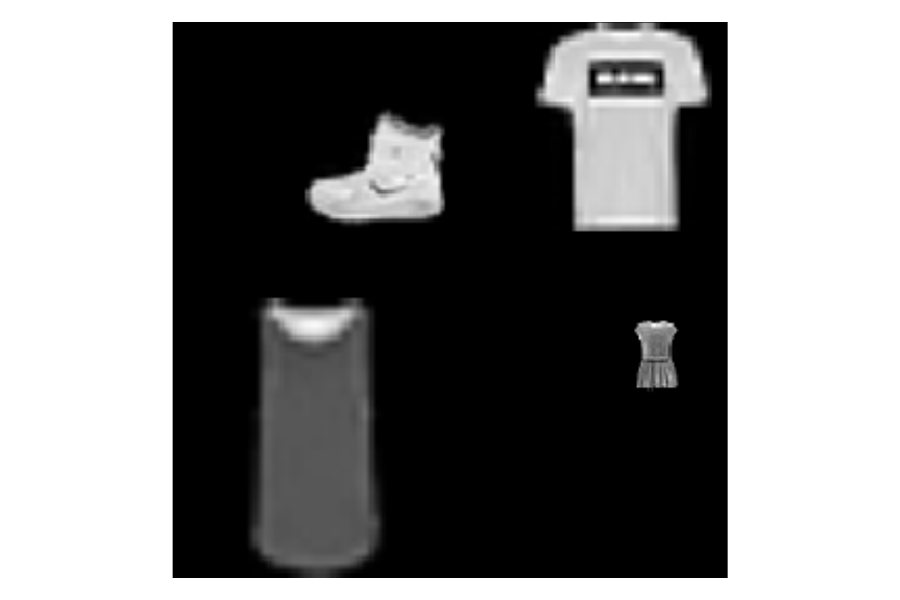} \\
      (a) Input image, homogeneous scales & (b) Input image, multi-scale \\
    \end{tabular}
  \caption{
    \textbf{Exp 1.1(B):}
    Example images from the \emph{multi-scale Fashion MNIST} toy dataset for segmentation. The corresponding segmentation masks are generated by assigning the class label of the corresponding object to pixels whose input grayscale values are above the threshold $h_{x,y} \geq h_\theta=0.2$.
  }
  \label{fig:FashionMNIST}
\end{figure}

\smallskip\noindent\textbf{Experiment~1.1(B): Segmentation of multi-scale inputs.}
We first evaluate the performance of \model~models on a small toy dataset, where each input image is formed by concatenating 4 images (objects) from the \emph{Fashion MNIST}~\cite{FMNIST} dataset (\fig{FashionMNIST}). 
Each object is assigned a random scale $s$, which determines the factor by which we upsample the original \emph{Fashion MNIST} image, via bilinear interpolation. 
The scale affects the object sizes --- the number of pixels occupied by the object in the image. 
We construct four different training sets: 
three where the scale of each object is homogeneous: the discrete variable $s$ has the probability mass functions $P(s=1)=1$, or $P(s=2)=1$, or $P(s=4)=1$; 
and one where the image contains objects on multiple scales: $s$ has the probability mass function $P(s)=0.25$ for values $s \in \{1,2,3,4\}$.
After rescaling, each object is placed in one quadrant of the input image, centered at a uniformly sampled random location. 
The corresponding ground truth segmentation masks are created by assigning the class label ($1 \ldots 10$) of the corresponding object to pixels whose input grayscale values $h_{x,y}$ are above the threshold $h_\theta=0.2$, by assigning the background label (0) to pixel locations where $h_{x,y}=0$, and by assigning an ignore index to undetermined pixel locations where $0<h_{x,y}<h_\theta$. 
The ignored pixel locations do not contribute to the loss during training and do not contribute to the accuracy at test time.

Due to the simple nature of the training set, we use a small \emph{U-Net} architecture, where the encoding network has three levels, as opposed to five in the original \emph{U-Net}~\cite{UNet}.
This corresponds to two downsampling layers. 
Each level is composed of two convolutional layers, followed by the ReLU activation layer. 
The channel dimension is 64 at the first level, and doubles with every downsampling, performed via $2 \times 2$ max pooling. 
In the decoding network, we use bilinear upsampling to increase feature map size, and in all convolutional layers we use `same' padding. We train all networks (\model~and baseline) for 50 epochs, using the ADAM optimizer~\cite{ADAMopt} and learning rate $0.0001$. 
To accommodate input images of different sizes, and keep with the original \emph{U-Net} implementation, we use a batch size of 1 and no batch normalization, but high momentum $\beta=(0.9, 0.999)$. 
To combat class imbalance, given the especially high frequency of the background class, we weigh the losses with the inverse of class frequencies in the training set. For the \model~models, we use filters with basis order 4 and 2. 
The scale parameter $\sigma$ is shared between all filters in a convolutional layer.

\begin{figure}[t]
\scriptsize
\centering
    \includegraphics[width=0.45\textwidth]{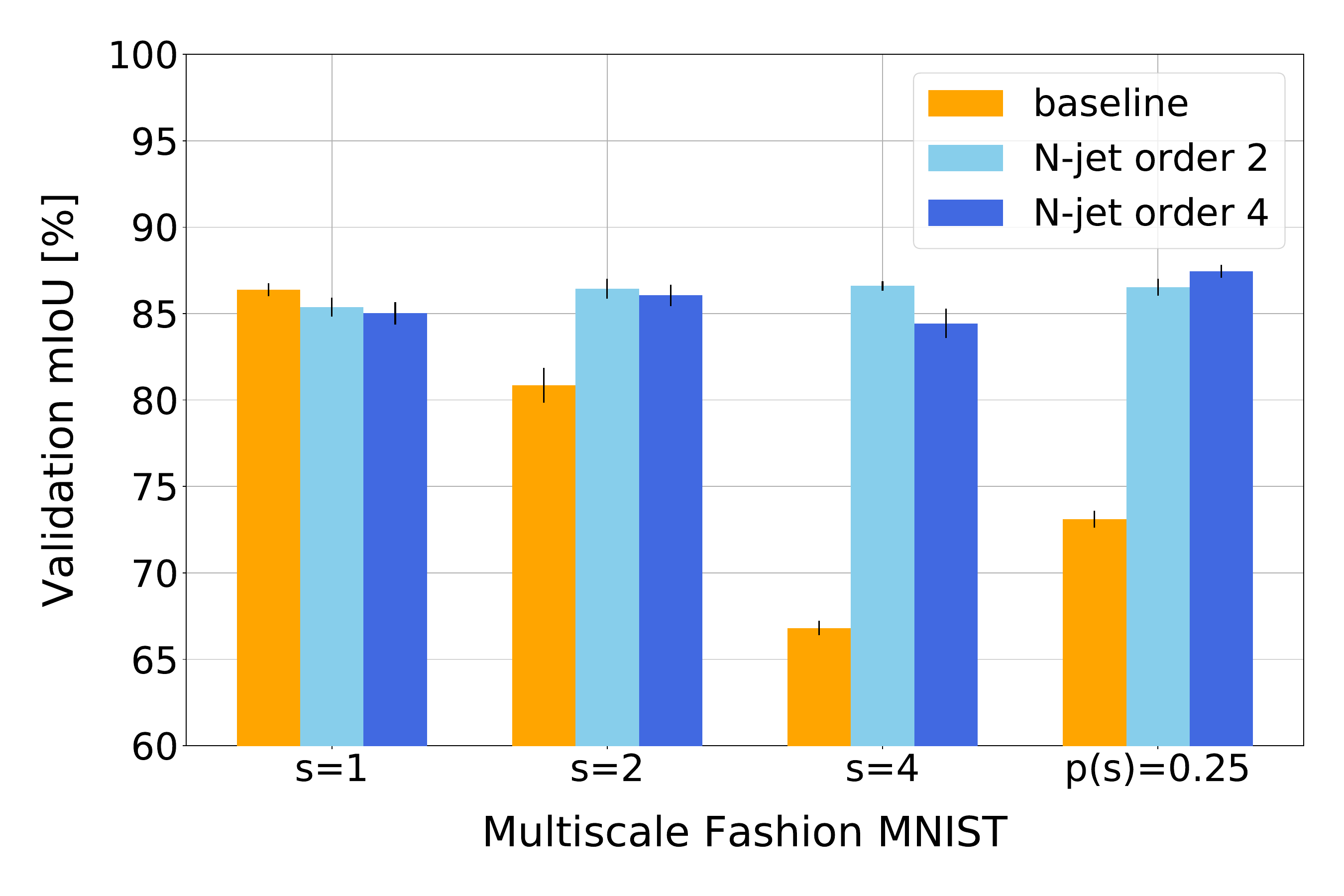}
  \caption{
  \textbf{Exp 1.1(B):}
    mIOU scores averaged over 5 repetitions with different random seeds on \emph{multi-scale Fashion MNIST}. Error bars denote the standard deviation. Validation mIoU decreases dramatically for baseline networks with fixed filter sizes (orange) for larger objects. In comparison, \model~models are robust against scale changes, as they can learn the filter size and scale, without changing the architecture or hyper-parameters.
  }
  \label{fig:Unet_mIoU}
\end{figure}

After training, we evaluate segmentation performance on the validation set using the mean intersection over union (mIoU) over all object classes. 
Each validation set is constructed in the same way as the corresponding training set, using the \emph{Fashion MNIST} validation images. We find that as we increase the average scale $s$ of the segmented objects (homogeneous scale case) or the variance of object scales $s$ (multi-scale case), \model~models successfully optimize the scale parameter $\sigma$ accordingly. This makes \model{s} capable of adapting to different object scales without changing the network architecture, depth or hyper-parameters at all. 
In contrast, baseline \emph{U-Net} models with fixed filter size cannot adapt their receptive field (RF) size based on the object scales in the training set, and their segmentation performance decays for larger objects (\fig{Unet_mIoU}).

In addition to the robustness of \model~networks against changing object scales, we find that \model{s} of only order 2 (where each kernel is defined by only 6 free parameters) is enough to obtain good validation accuracy. In fact, \model{s} of order 4 perform slightly worse for larger $s$. This is partly because the reduction of the basis order acts as a regularization via parameter reduction on our simple toy dataset, and partly because \emph{Fashion MNIST} (especially after upscaling) does not contain many high frequency components, which the higher order Gaussian derivatives can capture.

\begin{figure}[t]
\scriptsize
\centering
    \includegraphics[width=0.45\textwidth]{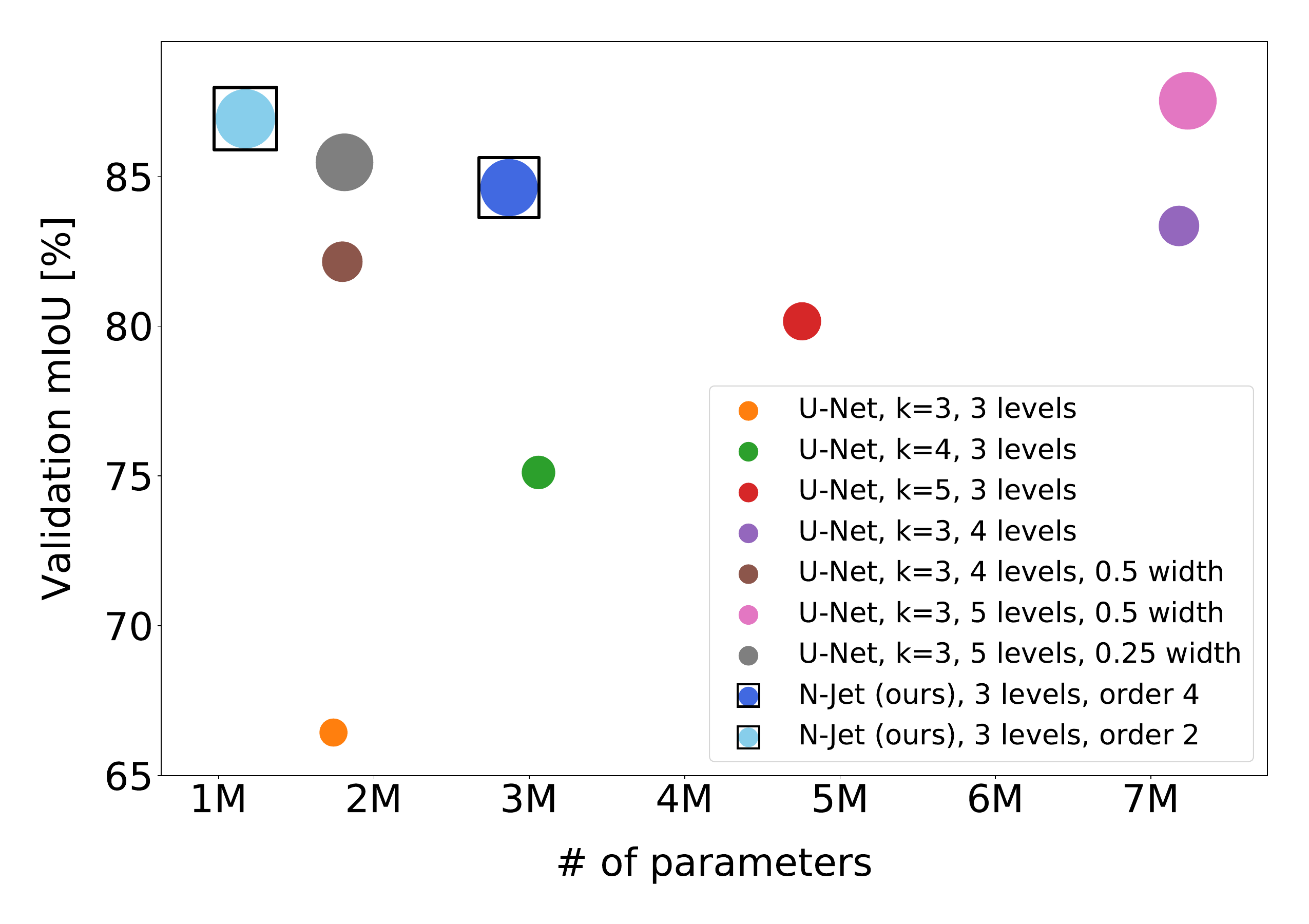}
  \caption{
    \textbf{Exp 1.2(B):}
    The effect of the number of parameters (x-axis) and receptive field size (round marker size) on the mIOU scores in the validation set of \emph{multi-scale Fashion MNIST} dataset with $s=4$.
    We observe that the number of model parameters is a weak predictor of segmentation performance compared to receptive field size which can be learned by \model{s} during training. We note that \model~models (in black squares) can perform dramatically better than the baseline model with the same architecture (orange circle), and overall display high mIoU with a low number of parameters.
  }
  \label{fig:Unet_no_params}
\end{figure}

\smallskip\noindent\textbf{Experiment 1.2(B): Learning the receptive field size.} 
While $\sigma$ optimization is successful for different basis orders, 
we note that the \model~model with basis order 4 has a larger number of free parameters than the baseline \emph{U-Net}. Nevertheless, we observe on our toy dataset that the validation mIoU depends only weakly on the number of parameters, beyond a certain network size. 
For the multi-scale segmentation task with $s=4$, where the scale of objects are increased by a factor of 4, we find that the receptive field size at the end of the encoding network largely determines the validation mIoU (\fig{Unet_no_params}). 
To demonstrate this, we vary the number of parameters and the receptive field size at the end of encoding in the baseline \emph{U-Net} models, until we match the \model~performance: we increase the kernel size $k$ from 3 to 4 and 5, and expand the depth of the baseline network by increasing the number of encoding and decoding levels from 3 (10 convolutional layers) to 4 (14 convolutional layers) and 5 (18 convolutional layers). 
To keep the number of trainable parameters at a reasonable level, for networks with 4 and 5 levels we also decrease the channel width of the layers (by halving or quartering the number of channels in each layer, as given in the legend of \fig{Unet_no_params}).

We find that the \model~models can outperform baseline \emph{U-Net} models while using a much smaller number of free parameters, due to $\sigma$ optimization.
In addition, we show that while the receptive field size is a good predictor of performance, it cannot be learned during training for the baseline \emph{U-Net}, and would need to be optimized via hyper-parameter scans. This can potentially mean increasing the depth of the network to match the input resolution, which cannot be parallelized.
Finally, we observe that slightly better validation mIoU can be obtained by baseline models, with almost 7 times the number of parameters and double the number of layers. We attribute this slight performance boost to the much larger depth, and thus increased number of nonlinearities in the network.

\medskip\noindent{\textbf{Experiment 2(B): Image segmentation using DeepLabv2.}}
Next, we consider a more realistic segmentation task on the \emph{Pascal VOC (SBD)} dataset~\cite{SBD,PascalVOC} using the \emph{DeepLabv2} architecture~\cite{chenPAMI17deeplab}. Modern \emph{DeepLab} models take advantage of dilated convolutions to aggregate information from multiple scales in atrous spatial pyramid pooling (ASPP) layers~\cite{chenPAMI17deeplab,Deeplabv3}. However, dilated kernels can only be upsampled discretely, based on the dilation rate in units of pixels. In addition, it is typically not possible to determine \emph{a priori} which scales in a dataset contain task-relevant information and the employed dilation rates need to be optimized using excessive hyper-parameter scans. We propose \model{s} as an alternative to optimizing the scales in a continuous way, eliminating the need to excessively search for dilation rates for each task.

To that end, we employ the \emph{DeepLabv2} model with a ResNet-101 backbone pretrained on the 20 class subset of the \emph{MS COCO} dataset~\cite{MSCOCO} corresponding to the \emph{Pascal VOC} classes.
We retain all the network and training hyper-parameters of the original \emph{DeepLabv2} model and finetune the baseline network with an ASPP output layer on \emph{Pascal VOC} with the batch normalization layers frozen.
For the \model~network, we replace the 4 convolutional layers of the ASPP layer with different dilation rates with 4 \model~layers with independent, learnable scales $\sigma$ (during finetuning) and we impose weight sharing between the different scales (i.e. same $\alpha$ values).
On top of eliminating the need to manually tune the dilation rates, \model~models with weight sharing also have the potential to dramatically reduce the number of parameters in ASPP layers.

\begin{table}[t]
    \caption{
       \textbf{Exp 2(B):}
       Segmentation mIoU scores on \emph{Pascal VOC} validation set along with the number of parameters in the ASPP layer. \model models with weight sharing, lower number of parameters, and no hyper-parameter tuning are as accurate as the baseline \emph{DeepLabv2} model with weight sharing.
       Moreover the \model nets substantially reduce the number of parameters.
    }
    \label{tab:deeplab_mIoU}
    \centering
    \normalsize

    \resizebox{1.0\linewidth}{!}{
    \begin{tabular}{ccc}
        \toprule
        \textbf{Model} & \textbf{mIoU} & \textbf{\# parameters} \\
        \midrule
        DeepLabv2 & 76.13 & 1,548,372 \\
        DeepLabv2, weight sharing & 74.73 & 387,093 \\
        \midrule
        \model, order 3 & 75.17 & 430,101 \\
        \model, order 2 & 74.58 & 258,069 \\
        \model, order 1 & 74.89 & 129,045 \\
        \bottomrule
    \end{tabular}
}
\end{table}
We find that \emph{DeepLabv2} with \model~output layers indeed allows for parameter reductions (\tab{deeplab_mIoU}). Using an \model~output layer with basis order 3 and weight sharing, we achieve validation mIoU values within 1\% of the baseline network, while reducing the number of parameters by nearly a factor of 4.
As an additional control, we also train a baseline network with weight sharing within the ASPP layer.
Interestingly, we observe that our \model~models attain on par or better performance than the weight-tied baseline network, even when we only use a basis order of 1 (each kernel is defined by only 3 free parameters). 

It is worth noting that these validation mIoUs are achieved with no hyper-parameter tuning for the \model~models, and despite not using \model~layers in the pretraining of the \emph{DeepLab} backbone. 
As it is, we believe \model~output layers may be used for multi-scale processing applications with further hyper-parameter tuning of learning rates and regularization parameters, or can be used out of the box to estimate the optimal scale or dilation rates for other architectures.

\section{Discussion and limitations}

\begin{figure}
    \centering
    \includegraphics[width=0.9\linewidth]{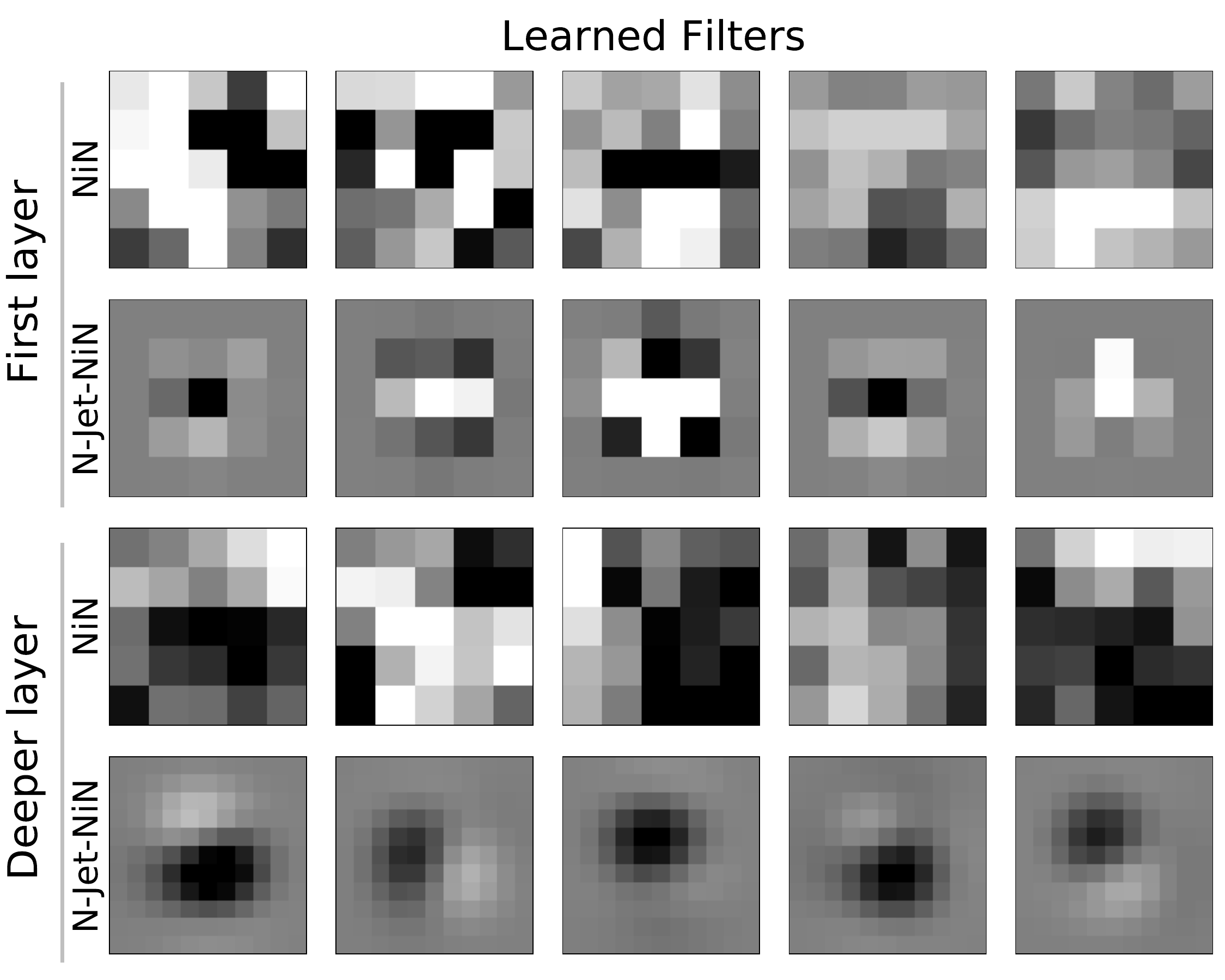}
    \caption{ 
    \textbf{Filter visualization.}
    Learned filters in the \emph{NiN} and \emph{\model-NiN} models. 
    In the first layer, \emph{\model-NiN} converges to a kernel size of $5 \times 5$, matching the baseline kernel size (top). In deeper layers, the learned $\sigma$ value can be much larger, leading to larger kernel sizes: in this example $11 \times 11$ (bottom). 
    All filter values are 0-centered.}
    \label{fig:srf_filters}
\end{figure}

To illustrate the differences between standard convolutional layers and \model~convolutional layers, we visualize a set of trained baseline filters compared to the equivalent \model filters (\fig{srf_filters}).
We find that in many models earlier layers will converge to smaller $\sigma$ values during training (\fig{srf_filters}, top), while deeper layers are prone to learning larger filter sizes (\fig{srf_filters}, bottom).

\begin{figure}
    \centering
    \includegraphics[width=0.8\linewidth]{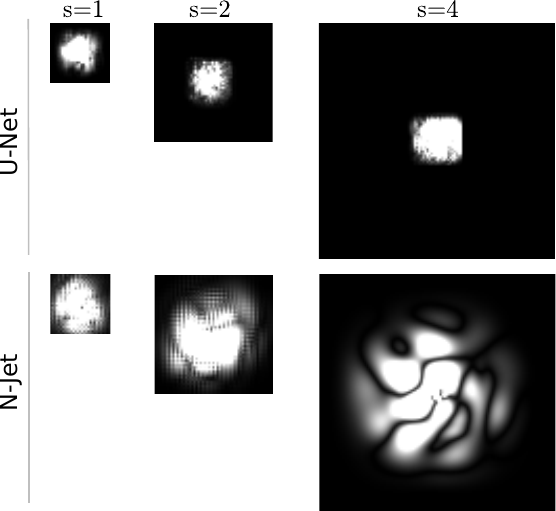}
    \caption{
    \textbf{Effective receptive field sizes.}
    Effective receptive field (eRF) size for the \model models trained on the multiscale Fashion-MNIST dataset. 
    Columns ($s=\{1, 2, 4\}$) denote models trained with input images scaled up by a factor of $s$.
    eRF visualizations are obtained by the gradients back-propagated to input pixel space~\cite{Luo2016understanding}. 
    We find that the eRF size of \model models scale up with the size of the input, while the eRF size of baseline U-Net models stay constant.}
    \label{fig:eRF_size}
\end{figure}

In addition, the strength of the \model representation lies in that it can learn filter sizes, and thus the receptive field size, during training. 
However, recent work has demonstrated that the effective receptive field (eRF) size of networks can be considerably smaller than what would be expected from the kernel size~\cite{Luo2016understanding}. 
We investigate the change in eRF size in our \model models by visualizing the gradients with respect to the input image in our models trained on the multiscale Fashion-MNIST dataset (\fig{eRF_size}).
We find that, as expected, the eRF size of \model models grows with the size of the training images, proportionally to the growth of filter sizes. 
The baseline U-Net model with $3 \times 3$ kernels cannot learn to adapt its receptive field size during training, its eRF size remains relatively constant as a function of the input image scale.

One of the limitations of our proposed kernels is that they are typically larger than the standard $3\times 3$ px, and therefore the convolutions take longer to compute.
This comes at no cost in parameters as the size of the \model filters is only affected by the scale parameters, $\sigma$. 
Additionally, computing the Gaussian basis is more expensive because it involves more operations: computing the Hermite polynomials, and obtaining the individual Gaussian basis from these, followed by estimating their linear combinations with the weights $\alpha$.
For the NiN architecture, our model is $\approx 2\times$ slower than the baseline model. 
As the network depth increases, so do the computations.  
However, manual architecture search takes a lot longer for finding the appropriate resolution hyper-parameters, because it requires a grid search over all possible filter sizes given a specific network depth and sub-sampling strategy.

\section{Conclusion}
We learn the resolution in deep convolutional networks. 
Learning the resolution frees the network architect from setting resolution related hyper-parameters such as the receptive field size and subsampling layers, which are dataset and network dependent. 
While we learn the receptive field size and the feature map subsampling for classification, the resolution is also determined by the depth of the network, as each layer increases the resolution linearly. Network depth is not something we learn, and thus we do not learn all resolution hyper-parameters. 
In addition to hard-coded filter sizes and subsampling layers, current CNN architectures are also designed to share the same filter size in a single layer. 
Due to computational restrains, our implementation does not make it possible to learn a $\sigma$ for each filter, rather than per layer. 
We leave this as potential future work.
To conclude, by replacing pixel-weights convolutional layers with our \layer layers we show that we can obtain similar performance as the baseline methods, 
without tuning the hyper-parameters controlling the resolution.

\smallskip\noindent\textbf{Acknowledgements.} 
This publication is part of the project "Pixel-free deep learning" (with project number 612. 001.805 of the research programme TOP which is financed by the Dutch Research Council (NWO).

{
    \bibliographystyle{ieeetr}
    \bibliography{srfBib}
}
\end{document}